\documentclass[preprint,12pt,authoryear]{elsarticle}

\usepackage{amssymb}
\usepackage{amsmath}

\usepackage{multirow}
\usepackage{graphicx}
\usepackage{url}
\usepackage{float}

\usepackage{hyperref}
\usepackage[round]{natbib}
\usepackage{xcolor}
\usepackage{microtype}

\journal{ISPRS Journal of Photogrammetry and Remote Sensing}

\begin{document}

\begin{frontmatter}



\title{A comprehensive and trustworthy benchmark of AI methods for change detection in Earth observation} 


\author[bvl,fmf]{Tadej Tomanič\corref{cor1}} 
\ead{tadej@bvlabs.ai}
\author[bvl]{Alice Baudhuin}
\author[bvl]{Jan Sotošek}
\author[bvl,ijs]{Jure Brence}
\author[bvl,ijs]{Panče Panov}
\author[bvl,ipp]{Nikola Simidjievski}
\author[bvl,ijs]{Dragi Kocev}

\cortext[cor1]{Corresponding author.}

\affiliation[bvl]{organization={Bias Variance Labs, d.o.o.},
            city={Ljubljana},
            postcode={1000}, 
            country={Slovenia}}

\affiliation[fmf]{organization={University of Ljubljana, Faculty of Mathematics and Physics},
            city={Ljubljana},
            postcode={1000}, 
            country={Slovenia}}

\affiliation[ijs]{organization={Department of Knowledge Technologies, Jo\v{z}ef Stefan Institute},
            city={Ljubljana},
            postcode={1000}, 
            country={Slovenia}}

\affiliation[ipp]{organization={Télécom Paris, Institut Polytechnique de Paris},
            city={Palaiseau},
            postcode={91128}, 
            country={France}}

\begin{abstract}

Change detection in Earth observation (EO) is critical for monitoring land surface transformations, yet  recent research in the field is constrained by inconsistent evaluation protocols and a narrow focus on predictive accuracy without regard for computational efficiency. To address this, we present a standardized, open-source benchmark for evaluating state-of-the-art (SOTA) deep learning methods for Earth observation change detection. We conduct a comprehensive analysis of ten representative model architectures, ranging from convolutional networks (CNNs) to vision transformers (ViTs), across ten heterogeneous change detection datasets. We rigorously evaluate these models with identical experimental protocols, comparing models trained from scratch against those utilizing pre-trained weights. Furthermore, we evaluate predictive performance alongside computational efficiency, including parameter counts and inference latency. Our findings reveal that well-optimized classical architectures, such as Siamese U-Nets, frequently outperform more complex contemporary models when computational efficiency is factored in, and that pre-training consistently provides a significant performance boost with no additional inference cost. To ensure complete transparency and reproducibility, \textit{all experimental resources}, including standardized data splits, training scripts, training logs, and model checkpoints are publicly available and adhere to FAIR principles (Findable, Accessible, Interoperable, and Reusable).

\end{abstract}

\begin{graphicalabstract}
\includegraphics[width=1\textwidth]{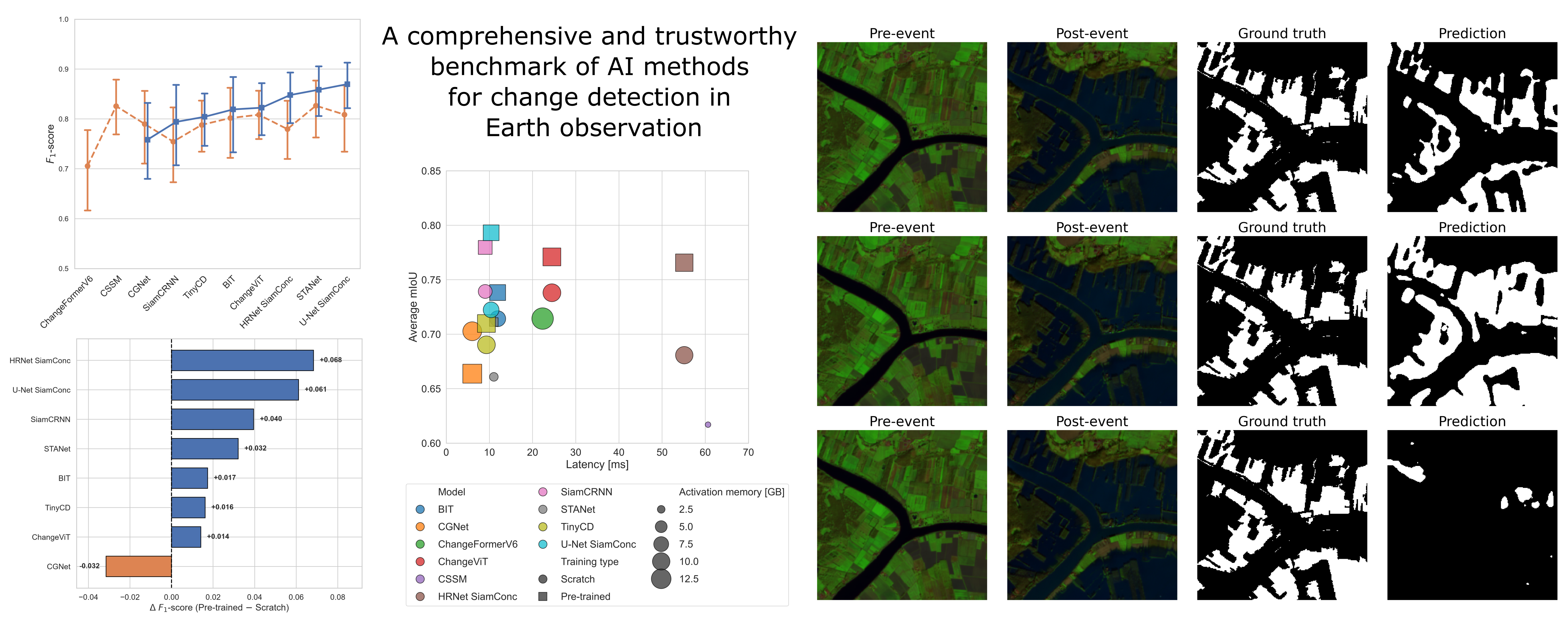}
\end{graphicalabstract}

\begin{highlights}
    \item Benchmarked 10 deep learning models across 10 diverse change detection datasets.
    \item Unified training and evaluation pipelines to minimize experimental bias.
    \item Analyzed trade-offs between model performance and computational efficiency.
    \item Openly released all resources in strict adherence to FAIR principles.
\end{highlights}

\begin{keyword}
Change detection \sep Benchmarking study \sep Computer vision \sep Trustworthy AI \sep FAIR principles



\end{keyword}

\end{frontmatter}


\section{Introduction}
\label{sec:intro}

Change detection (CD) lies at the heart of modern Earth observation (EO), enabling automated monitoring of land surface transformations across space and time. At its core, change detection is the process of identifying semantic differences in a specific geographical area by analyzing images captured at different times. This typically involves comparing co-registered bi-temporal image pairs to distinguish genuine structural or land-cover transformations from irrelevant variations, such as seasonal shifts \citep{Shafique2022, Cheng2024, Yang2026a}.

Driven by the rapid expansion of modern satellite constellations, ranging from high-resolution optical platforms to multi-spectral sensors (e.g., Sentinel-2) and all-weather Synthetic Aperture Radar (SAR) platforms (e.g., Sentinel-1), multi-temporal imagery is acquired continuously at unprecedented spatial, spectral, and temporal resolutions \citep{Yang2026a, Jabbarizadegan2026}. The ability to rapidly analyze physical changes between temporal acquisitions supports critical real-world applications, including urban growth tracking, agricultural monitoring, post-disaster damage mapping, and environmental conservation \citep{Yang2026a, Jabbarizadegan2026}. As the volume and complexity of multi-temporal satellite data scale exponentially, automated deep learning models have become vital for detecting meaningful changes across land surfaces \citep{Yang2026a, Jabbarizadegan2026}.

However, the field of EO change detection suffers from empirical fragmentation caused by a rapid rise and ad of new models and inconsistent evaluation protocols \citep{Corley2024}. Recent literature has seen a rapid shift from convolutional neural networks (CNNs) \citep{Daudt2018, Sun2019, Chen2020b} to vision transformers (ViTs) \citep{Chen2022, Bandara2022, Zhu2026} and state-space models (SSMs) \citep{Ghazaei2026}, with each new architecture claiming superior performance over established baselines \citep{Corley2024}. Regrettably, these claims are frequently built on inconsistent evaluation setups. Comparative studies in the literature routinely use non-standardized train/validation/test splits, varying image patch resolutions, non-uniform loss functions, and disparate hyperparameter configurations \citep{Corley2024}. Crucially, fair comparisons and auditing reveals that when evaluated within strictly controlled, identical training pipelines, traditional models established nearly a decade ago \citep{Daudt2018}, frequently match or outperform complex modern state-of-the-art (SOTA) architectures \citep{Corley2024}. Furthermore, mainstream literature heavily emphasizes minor accuracy gains while ignoring computational overhead, omitting essential reporting on floating-point operations (FLOPs), trainable parameter counts, memory footprints, and other metrics crucial for operational deployment on edge platforms or large-scale processing pipelines \citep{Yang2026b}.

To eliminate protocol bias and establish a reliable baseline for model comparison, a unified and open-source benchmark is required. A rigorous change detection benchmark must evaluate diverse model architectures with strictly identical experimental protocols, enforcing uniform dataset splits, unified data processing, standardized loss functions, and identical performance metrics \citep{Corley2024}. Furthermore, to be truly representative of modern EO challenges, such a benchmark must evaluate architectures across heterogeneous datasets, encompassing urban development, seasonal vegetation shifts, post-disaster damage, and multi-sensor acquisitions, while simultaneously profiling the trade-off between performance and computational efficiency \citep{Yang2026b}.

Beyond controlled comparison, a benchmark is only useful to the extent that its results can be \textit{trusted}. Recent trustworthy artificial intelligence (AI) frameworks \citep{HLEG2019, NIST2023,Dimitrovski2023,DIMITROVSKI202318} identify transparency, reliability, and accountability as core requirements for AI systems, and these translate directly into requirements for the benchmarks used to evaluate them. We therefore treat trustworthy benchmarking as resting on four pillars: (i) \textit{methodological fairness}, achieved through strictly identical training and evaluation pipelines that isolate architectural effects from confounding protocol differences; (ii) \textit{transparency and reproducibility}, achieved by fixing dataset splits, logging complete experiment metadata, and openly releasing all code, checkpoints, and logs; (iii) \textit{reliability}, achieved by evaluating across heterogeneous datasets and reporting performance variance rather than single-dataset optima; and (iv) \textit{FAIRness} of the released artifacts, ensuring that datasets, software, and models are Findable, Accessible, Interoperable, and Reusable (FAIR) \citep{Wilkinson2016}. These four pillars organize the benchmark design described in the remainder of this paper.

Concretely, pillar~(i) is operationalized by the unified training and evaluation protocol of Section~\ref{sub:traineval}; pillar~(ii) by the fixed, released splits and complete experiment logging described in Section~\ref{sub:repro}; pillar~(iii) by the cross-dataset evaluation of Section~\ref{sub:performance}; and pillar~(iv) by the FAIR artifact release and ontological annotation of Section~\ref{sub:repro}. We return to these four pillars in Section~\ref{sec:conclusions}.

To address these requirements, we introduce a novel standardized benchmark for Earth observation change detection. This work has several main contributions. Specifically, we:

\begin{itemize}
    \item Establish a strictly controlled experimental protocol featuring fixed, reproducible dataset splits, unified data processing, training and evaluation pipelines, and standardized loss functions, minimizing setup discrepancies and evaluation bias;
    \item Perform a comprehensive evaluation of textbf{ten} representative AI methods for change detection across \textbf{ten} heterogeneous change detection datasets, covering diverse scenarios including urban development, land-cover changes, and post-disaster damage;
    \item Derive study-design principles for training and evaluating AI methods for change detection and discuss common practical aspects that affect their application and performance in the context of these tasks.   
    \item Provide a fully open-source benchmarking framework integrated into the AiTLAS toolbox \citep{Dimitrovski2023, AiTLAS2026}, releasing all standardized data splits, data loaders, training scripts, schemas and ontology on GitHub, alongside trained model checkpoints and TensorBoard logs on Hugging Face and Zenodo to ensure end-to-end reproducibility.
\end{itemize}

\section{Related work}
\label{sec:related}

Several papers have reviewed the growing literature on deep learning for change detection in Earth observation \citep{Shafique2022, Cheng2024, Gomroki2026, Jabbarizadegan2026, Yang2026a}. 

To begin with, \citet{Shafique2022} surveyed deep learning-based change detection across diverse modalities, including SAR, multispectral, hyperspectral, Very-High-Resolution (VHR), and heterogeneous imagery. They categorized methods by their supervision paradigms (supervised, unsupervised, and semi-supervised) and highlighted critical operational bottlenecks, including labeled data scarcity, registration noise, scene complexity, and the literature's predominant focus on binary over multiclass change detection.

Furthermore, \citet{Cheng2024} provided a comprehensive review of EO change detection over the past decade. They categorize existing literature across three core dimensions: algorithm granularity (pixel, region, and hybrid), supervision modes, and learning frameworks (ranging from traditional models and CNNs to ViTs and SSMs) while benchmarking SOTA performance across multimodal datasets (SAR, multispectral, hyperspectral, heterogeneous, and 3D).

A recent case study on the evolution of optical change detection from Siamese CNNs to ViTs and SSMs emphasized that real-world deployment requires task-specific pipelines to handle practical constraints like registration errors, atmospheric noise, and class imbalance \citep{Yang2026a}.

Similarly, a recent systematic review reports on 144 studies across statistical, machine learning, and deep learning methods for change detection in EO imagery. Crucially, their meta-analysis reveals that architectural performance gains have saturated on standard benchmarks, demonstrating that operational progress is now constrained more by annotation costs, label availability, and domain shift than by model capacity \citep{Jabbarizadegan2026}.

Finally, a recent survey traces the transition from pixel- and object-based methods for deep learning across 2D and 3D data (e.g., LiDAR, point clouds) for urban change detection. The authors emphasize that while 2D change detection is mature, 3D deep learning remains in its infancy, mostly constrained by high acquisition costs, data noise, and computational intensity \citep{Gomroki2026}.

\subsection{Reality check}
\label{sub:reality_check}

The need for controlled change detection benchmarking have been well motivated by \citet{Corley2024}. Investigating whether the rapid proliferation of task-specific architectures truly represents genuine progress, the authors demonstrated that standard baseline models, such as U-Net models that build on ImageNet-pretrained backbones (e.g., ResNet-50) can consistently match or outperform complex transformer-based models like BIT \citep{Chen2022} and ChangeFormer \citep{Bandara2022} across canonical benchmarks, such as LEVIR-CD \citep{Chen2020a} and WHU-CD \citep{Ji2019}.

Their study revealed that published claims of SOTA gains frequently stem from three widespread evaluation biases. Firstly, new architectures are often compared against outdated literature figures rather than re-evaluated under unified optimization setups. For instance, early baselines such as fully-convolutional networks were hardcoded without pretrained backbones, making direct comparisons against modern, pretrained models inherently unfair \citep{Daudt2018}. Secondly, widespread reliance on buggy preprocessed datasets—most notably a popular random split of WHU-CD that inadvertently leaked roughly 85\% of test patches into the training set—inflated scores and rendered published cross-paper comparisons invalid \citep{Ji2019}. When evaluated on official, uncorrupted splits, complex models suffered severe performance drops. Finally, model performance improvements are routinely misattributed to novel modules when they actually result from modernized training procedures, such as updated loss functions, learning rate schedules, or data augmentations.

The main findings from the reality check by \citet{Corley2024} suggest the primary bottleneck in change detection research is no longer model capacity, but evaluation bias and non-standardized experimental procedures. Ensuring that performance gains reflect genuine architectural progress requires moving away from isolated model proposals toward a strictly controlled benchmark. Therefore, Section~\ref{sec:data_models} details the datasets and models selected for our benchmark, while Section~\ref{sec:experiments} outlines our experimental design, including training and evaluation protocol, performance metrics, and experiment tracking and reproducibility.

\subsection{Trustworthy and reproducible benchmarking}\label{sub:trustworthy}

The requirements identified by trustworthy AI frameworks \citep{HLEG2019, NIST2023}---transparency, reliability and accountability---are typically formulated for deployed AI systems, but they apply with equal force to the benchmarks on which such systems are selected: a model chosen on the basis of an unauditable comparison inherits that opacity. 

The evaluation biases documented for change detection are a domain-specific instance of a broader reproducibility problem in machine learning \citep{Pineau2021}, where reported gains frequently fail to survive re-evaluation under standardized conditions. In response, the community has converged on practices that make experimental claims auditable and reusable: standardized reporting of models and datasets through model cards \citep{Mitchell2019} and datasheets \citep{Gebru2021}, and the application of the FAIR guiding principles \citep{Wilkinson2016} not only to data but also to research software \citep{Barker2022} and to AI models and pipelines \citep{Huerta2023}. Relatedly, \citet{Bouthillier2021} demonstrated that a substantial share of reported improvements in machine learning benchmarks falls within the variance induced by sources of randomness that single-run protocols leave unmeasured, which motivates reporting the spread of results alongside their central tendency. 

Together these establish that trustworthy benchmarking requires more than a fair comparison at a single point in time; it requires that the datasets, code, and trained models underlying a result remain discoverable, accessible, and reusable by others. Our benchmark operationalizes these principles for EO change detection, combining a controlled comparison protocol (Section~\ref{sec:experiments}) with a FAIR, openly released set of artifacts (Section~\ref{sub:repro}).

\section{Datasets \& models}
\label{sec:data_models}

\subsection{Benchmarking datasets}
\label{sub:data}

To rigorously benchmark model generalization under diverse real-world conditions, we evaluate model performance across ten widely used Earth observation change detection datasets. These datasets were selected to encompass a broad range of spatial resolutions, sensor modalities, geographic regions, and scene dynamics. We categorize these datasets into the following thematic categories:

\begin{itemize}
    \item \textbf{Urban and infrastructure monitoring:} LEVIR-CD+ \citep{Chen2020a}, EGY-BCD \citep{Holail2023}, and BANDON \citep{Pang2023} focus on anthropogenic structural dynamics, primarily building construction and demolition. Notably, BANDON incorporates off-nadir aerial imagery to address the challenging scenario of oblique viewpoints and parallax distortions.
    \item \textbf{Environmental and disaster mapping:} OMBRIA \citep{Drakonakis2022} targets sudden disaster-induced surface transformations, focusing on post-disaster flood inundation mapping.
    \item \textbf{Agricultural and land cover dynamics:} CLCD \citep{Liu2022} tracks spatio-temporal dynamics in agricultural landscapes, isolating true cropland conversions from background soil variations and seasonal crop growth.
    \item \textbf{General and multi-sensor change detection:} Season-Varying CDD \citep{Lebedev2018}, DSIFN \citep{Zhang2020}, SYSU-CD \mbox{\citep{Shi2022}}, MSBC and MSOSCD \citep{Li2022} capture multi-class surface changes across diverse urban and rural terrains. MSBC and MSOSCD specifically integrate multi-source observations spanning optical multispectral, SAR, and VHR sensors.
\end{itemize}

Note that none of these datasets have been used for pre-training but solely for benchmarking. The benchmarked datasets introduce three core challenges essential for evaluating EO models. Firstly, the datasets span Ground Sample Distances (GSD) from sub-meter aerial imagery ($0.03$--$0.6$~m/px in LEVIR-CD+ \citep{Chen2020a}, SYSU-CD \citep{Shi2022}, EGY-BCD \citep{Holail2023} and BANDON \citep{Pang2023}) to medium-resolution spaceborne platforms ($10$~m/px in Sentinel-based datasets like OMBRIA \citep{Drakonakis2022}). Furthermore, multi-sensor inputs (e.g., Sentinel-1 SAR combined with Sentinel-2 optical imagery in OMBRIA, MSBC and MSOSCD \citep{Li2022}) test model resilience against geometric misregistration and SAR speckle noise.

Secondly, unlike ideal nadir acquisitions, off-nadir imagery (such as in BANDON \citep{Pang2023}) introduces severe building side-wall relief displacement. Models must distinguish true ground-level structural changes from visual displacement artifacts caused by non-zero zenith angles.

Thirdly, seasonal vegetation shifts, soil moisture variation, and sun-angle illumination differences introduce high intra-class variance. Datasets such as Season-Varying CDD \citep{Lebedev2018} and CLCD \citep{Liu2022} explicitly require algorithms to suppress environmental phenological shifts while detecting true semantic changes.

The key information, such as image characteristics, spatial resolutions, and primary tasks of all ten benchmarking datasets are summarized in Table~\ref{tab:datasets}.

\begin{table}[htbp]
    \centering
    \caption{Key characteristics of the change detection datasets used for the benchmarking experiments.}
    \vspace{5pt}
    \small
    \resizebox{\textwidth}{!}{
    \begin{tabular}{|l|l|c|c|c|c|}
    \hline
    \textbf{Dataset} & \textbf{Image source} & \textbf{Image pairs} & \textbf{Bands} & \textbf{Patch size} & \begin{tabular}[c]{@{}c@{}}\textbf{GSD/} \\ \textbf{Resolution}\end{tabular} \\
    \hline
    BANDON \citep{Pang2023} & Off-nadir aerial & 2{,}283 & 3 & $2048 \times 2048$ & $0.6$~m \\
    \hline
    CLCD \citep{Liu2022} & GaoFen-2 & 600 & 3 & $512 \times 512$ & $0.5$--$2$~m \\
    \hline
    DSIFN \citep{Zhang2020} & Google Earth & 3{,}988 & 3 & $512 \times 512$ & $0.5$--$2$~m \\
    \hline
    EGY-BCD \citep{Holail2023} & Google Earth & 6{,}091 & 3 & $256 \times 256$ & $0.25$~m \\
    \hline
    LEVIR-CD+ \citep{Chen2020a} & Google Earth & 985 & 3 & $1024 \times 1024$ & $0.5$~m \\
    \hline
    \multirow{3}{*}{MSBC \citep{Li2022}} & GaoFen-2 & \multirow{3}{*}{3{,}769} & 3 & \multirow{3}{*}{$256 \times 256$} & \multirow{3}{*}{$0.8$--$10$~m} \\
     & Sentinel-2 & & 7 & & \\
     & Sentinel-1 & & 2 & & \\
    \hline
    \multirow{3}{*}{MSOSCD \citep{Li2022}} & GaoFen-2 & \multirow{3}{*}{5{,}107} & 3 & \multirow{3}{*}{$256 \times 256$} & \multirow{3}{*}{$0.8$--$10$~m} \\
     & Sentinel-2 & & 7 & & \\
     & Sentinel-1 & & 2 & & \\
    \hline
    \multirow{2}{*}{OMBRIA \citep{Drakonakis2022}} & Sentinel-1 & \multirow{2}{*}{694} & 1 & \multirow{2}{*}{$256 \times 256$} & \multirow{2}{*}{$10$~m} \\
     & Sentinel-2 & & 3 & & \\
    \hline
    Season-Varying CDD \citep{Lebedev2018} & Google Earth & 16{,}000 & 3 & $256 \times 256$ & $0.03$--$1$~m \\
    \hline
    SYSU-CD \citep{Shi2022} & Aerial & 20{,}000 & 3 & $256 \times 256$ & $0.5$~m \\
    \hline
    \end{tabular}
    }
    \label{tab:datasets}
\end{table}

\begin{figure}[htbp]
    \centering
    \includegraphics[width=1\textwidth]{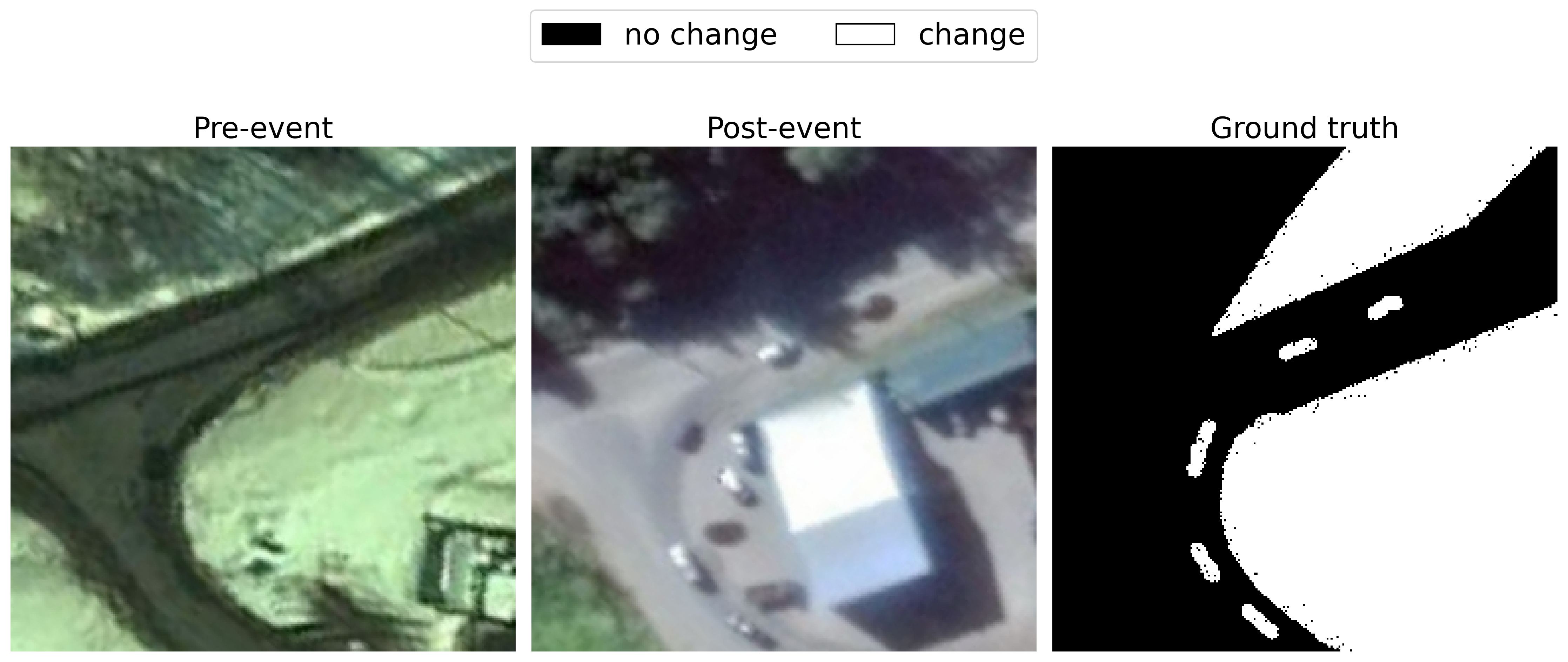}
    \caption{Example image pair ($T_1, T_2$) and ground truth change mask from the Season-Varying CDD dataset \citep{Lebedev2018}.}
    \label{fig:cd_example}
\end{figure}

Each dataset provides co-registered pre-event ($T_1$) and post-event ($T_2$) image pairs accompanied by binary pixel-level ground truth masks $Y \in \{0, 1\}^{H \times W}$, where $1$ denotes a \textit{change} event and $0$ represents \textit{no change} background pixels. An example of bi-temporal image pair and its corresponding ground truth mask from the Season-Varying CDD dataset \citep{Lebedev2018} are depicted in Figure~\ref{fig:cd_example}.

\subsection{Model architectures}
\label{sub:models}

To evaluate performance across diverse model architectures, we benchmark ten representative change detection models spanning classical convolutional networks, attention and prior-guided networks, vision transformers, and recent state-space models.

An overview of the benchmarked change detection models, including their parameter counts, and theoretical computational overhead (see Subsection \ref{subsub:compute_efficiency} for details), is provided in Table~\ref{tab:models}.

\begin{table}[htbp]
    \centering
    \caption{Summary of the representative model architectures considered in this benchmark. FLOPs and parameter counts are reported for bi-temporal input tensor pairs of size $2 \times 3 \times 256 \times 256$.}
    \vspace{5pt}
    \small
    \resizebox{\textwidth}{!}{
    \begin{tabular}{|l|c|c|c|}
    \hline
    \textbf{Model} & \textbf{Year} & \textbf{Params (M)} & \textbf{GFLOPs} \\
    \hline
    BIT \citep{Chen2022} & 2021 & 3.50 & 10.88 \\
    \hline
    CGNet \citep{Han2023} & 2023 & 33.68 & 87.55 \\
    \hline
    ChangeFormerV6 \citep{Bandara2022} & 2022 & 41.03 & 138.77 \\
    \hline
    ChangeViT \citep{Zhu2026} & 2026 & 20.66 & 26.36 \\
    \hline
    CSSM \citep{Ghazaei2026} & 2025 & 2.59 & 2.78 \\
    \hline
    HRNet SiamConc \citep{Sun2019} & 2019 & 21.48 & 9.55 \\
    \hline
    SiamCRNN \citep{Chen2020b} & 2020 & 28.51 & 65.49 \\
    \hline
    STANet \citep{Chen2020a} & 2020 & 12.21 & 19.15 \\
    \hline
    TinyCD \citep{Codegoni2023} & 2023 & 0.29 & 1.46 \\
    \hline
    U-Net SiamConc \citep{Daudt2018} & 2018 & 40.35 & 19.37 \\
    \hline
    \end{tabular}
    }
    \label{tab:models}
\end{table}

\subsubsection{Convolutional networks}

Convolutional networks establish baseline performance by extracting spatially localized features. The following such models were used in our benchmark:

\begin{itemize}
    \item \textbf{U-Net SiamConc} \citep{Daudt2018}: A canonical fully convolutional Siamese U-Net. Bitemporal feature maps extracted by parallel branches are concatenated across skip connections prior to decoder propagation, setting a standard baseline for bi-temporal segmentation.
    \item \textbf{HRNet SiamConc} \citep{Sun2019}: Maintains high-resolution representations throughout feature extraction by interconnecting multi-resolution convolutions in parallel. This prevents spatial information loss along deep layers, ensuring precise boundary localization.
    \item \textbf{SiamCRNN} \citep{Chen2020b}: A hybrid architecture combining Siamese convolutional feature extractors with recurrent neural networks (RNNs) to jointly model spatial structures and complex temporal-spectral transitions across acquisitions.
    \item \textbf{TinyCD} \citep{Codegoni2023}: An ultra-lightweight, resource-aware architecture utilizing a mix and attention mask block (MAMB) to process spatio-temporal features, optimized for edge-computing constraints while maintaining competitive accuracy at a fraction of the computational complexity.
\end{itemize}

\subsubsection{Attention and prior-guided networks}

To tackle scene complexity, illumination shifts and boundary degradation, specialized attention mechanisms and prior maps refine convolutional models. These models include:

\begin{itemize}
    \item \textbf{STANet} \citep{Chen2020a}: Integrates spatial-temporal self-attention modules (pyramid attention module) within a Siamese metric-learning framework to capture long-range spatial dependencies and improve invariance against environmental variation.
    \item \textbf{CGNet} \citep{Han2023}: Incorporates an explicit deep change prior map to guide multi-scale feature fusion via a change guide module (CGM), effectively suppressing boundary errors and internal hole artifacts in changed regions.
\end{itemize}

\subsubsection{Vision transformers}

Transformer-based models capture global contextual interactions and long-range dependencies across bi-temporal pairs. The list of models in this benchmark includes:

\begin{itemize}
    \item \textbf{BIT} \citep{Chen2022}: Expresses bi-temporal spatial features as compact semantic tokens, employing a transformer encoder-decoder to model spatial-temporal context in token space before projecting predictions back to pixel space.
    \item \textbf{ChangeFormerV6} \citep{Bandara2022}: A unified, purely transformer-based Siamese network combining a hierarchical transformer encoder with a lightweight multi-layer perceptron (MLP) decoder to extract multi-scale context without convolutional operations.
    \item \textbf{ChangeViT} \citep{Zhu2026}: Adapts a plain vision transformer backbone paired with a detail-capture module and feature injector, balancing global context extraction with fine-grained edge preservation.
\end{itemize}

\subsubsection{State-space models}

SSMs leverage linear-time sequence modeling and aim to achieve high computational efficiency while maintaining global receptive fields. The model used as part of the benchmark includes:

\begin{itemize}
    \item \textbf{CSSM} \citep{Ghazaei2026}: A selective state-space (Mamba-based) architecture designed specifically for change detection. It employs target-specific $L_1$ parameter selection to highlight relevant changed features while discarding pseudo-changes with linear computational complexity.
\end{itemize}

\section{Experimental design}
\label{sec:experiments}

\subsection{Training and evaluation protocol}
\label{sub:traineval}

\subsubsection{Data splits}

A $60/20/20$ train/validation/test split was applied across the datasets used in the benchmarking experiments. Season-Varying CDD, SYSU-CD, and CLCD datasets already provide such a split, as defined by the original authors of the datasets, and these were used as-is. For all other remaining datasets (LEVIR-CD+, DSIFN, MSBC, MSOSCD, EGY-BCD, BANDON, and OMBRIA), we generated the splits by random sampling.

\subsubsection{Data preprocessing}

To ensure consistent image dimensions and value scaling across diverse sensors and spectral configurations, custom data transform pipelines were applied.

Input images were preprocessed using per-channel min-max normalization to rescale feature values into $[0, 1]$:

\begin{equation}
    \tilde{X}_c = \frac{X_c - \min(X_c)}{\max(X_c) - \min(X_c) + \epsilon},
    \label{eq:minmax}
\end{equation}

\noindent where $X_c$ denotes the $c$-th channel matrix, and $\epsilon = 10^{-7}$ prevents numerical instability during division. Channels exhibiting zero variance (where $\max = \min$) were explicitly zeroed out. Normalized inputs were permuted from channels-last $(H, W, C)$ to channels-first $(C, H, W)$ tensors and spatially resized to $256 \times 256$ with antialiasing enabled.

Similarly, ground truth change maps were transformed into $(C, H, W)$ tensors, cast to 32-bit floating-point tensors, and resized to match the $256 \times 256$ spatial target resolution.

\subsubsection{Training and optimization strategy}

All benchmarking models were implemented and trained using the AiTLAS toolbox \citep{Dimitrovski2023, AiTLAS2026}. We used a maximum of 100 epochs with a batch size of 16 and an initial learning rate of $1 \times 10^{-4}$. Data loading was executed using 4 worker threads. Training data was shuffled randomly at each epoch, whereas validation and test data ordering was kept deterministic. The models were optimized using a hybrid loss function comprising an equal weighting ($w_1 = w_2 = 0.5$) of focal loss \citep{Lin2018} and dice loss \citep{Milletari2016}:

\begin{equation}
    \begin{aligned}
    \mathcal{L} = w_1 \left[ - \frac{1}{N} \sum_{i=1}^{N} \alpha_t (1-p_{t,i})^{\gamma} \log (p_{t,i}) \right] + \\ w_2 \left[ 1 - \frac{2 \sum_{i=1}^{N} p_i y_i + \epsilon}{\sum_{i=1}^{N} p_i + \sum_{i=1}^{N} y_i + \epsilon} \right],
    \end{aligned}
    \label{eq:hybrid_loss}
\end{equation}

\noindent where $N$ is the total number of pixels in the image, $y_i \in \{0, 1\}$ is the ground-truth label for the $i$-th pixel, and $p_i \in [0, 1]$ is the predicted probability for the positive class. For the focal loss term, $p_{t,i}$ denotes the predicted probability of the true class (defined as $p_{t,i} = p_i$ if $y_i = 1$, and $1-p_i$ if $y_i = 0$), and $\alpha_t$ is the corresponding weighting factor ($\alpha_t = \alpha$ if $y_i = 1$, and $1-\alpha$ if $y_i = 0$). In our experiments, we set the focusing parameter $\gamma = 2$, the class balancing weight $\alpha = 0.25$, and the loss combination weights $w_1 = 0.5$ and $w_2 = 0.5$.

To evaluate the influence of domain transfer, all architectures were trained and evaluated under two distinct initialization regimes:

\begin{itemize}
    \item Random initialization: Models trained completely from scratch with randomly initialized weights.
    \item Transfer learning: Models initialized with weights pre-trained on ImageNet-1K.
\end{itemize}

The only exception are ChangeFormerV6 and CSSM, which did not have pre-trained weights available and were trained from scratch only. Note that pre-trained weights were applied strictly to the model backbones, as full end-to-end pre-trained checkpoints for these model architectures are generally not available. All remaining architectural modules, such as fusion modules or prediction heads, were initialized randomly using standard schemes. The majority of backbone weights, pre-trained on the Image-Net-1K dataset, were obtained from the \textit{torchvision} model repository \citep{Torchvision2016}. As an exception, ChangeViT utilizes a hybrid encoder that combines both ResNet18 and data-efficient image transformer (DeiT) architectures. The ResNet18 weights were sourced from \textit{torchvision}, whereas the DeiT weights were obtained directly from the original repository.

To prevent overfitting, early stopping after 10 epochs was implemented on the validation split for each dataset. The model checkpoint exhibiting the lowest validation loss was saved and subsequently evaluated on the test split to determine its final predictive performance.

\subsection{Evaluation metrics}
\label{sub:metrics}

To evaluate candidate models comprehensively, we measure both predictive performance and computational efficiency.

\subsubsection{Predictive performance}
\label{subsub:performance}

During the testing phase, continuous output probability maps generated by the networks were discretized into binary change detection masks using a fixed decision threshold of $\tau = 0.5$.

Model generalization was evaluated on the unseen test split. Because change detection datasets typically suffer from severe spatial class imbalance—where changed pixels represent a small fraction of the total area—overall accuracy alone can be misleading. Therefore, mIoU and $F_1$-score  serve as the primary benchmarking metrics to rigorously evaluate model performance.

\subsubsection{Computational efficiency}
\label{subsub:compute_efficiency}

To profile operational feasibility for real-world deployment, we evaluate predictive performance directly against computational efficiency metrics, including parameter counts, floating operations (FLOPs), latency/inference time, activation memory, and training time. Model capacity is represented by the total number of trainable parameters in millions ($\text{M}$) and computational complexity ($\text{FLOPs}$) is estimated for a single forward pass on a standardized $2 \times 3 \times 256 \times 256$ bi-temporal input tensor. Meanwhile, latency ($\text{ms}$) records average wall-clock execution time per image pair on a single GPU with a batch size of 1. To evaluate optimization overhead, peak activation memory ($\text{MB}$) is measured across a standard training batch size of 16, and total training time ($\text{min}$) tracks the wall-clock duration required for the network to reach convergence under our early-stopping protocol.

\subsection{Reproducibility, FAIR release, and trustworthy integration}
\label{sub:repro}

To ensure full transparency, traceability, and experimental reproducibility, all training runs were monitored and structured using MLflow \citep{Zaharia2018}. For every benchmarking experiment, the MLflow tracking server logged complete execution metadata, dataset split identifiers, and hyperparameter settings. Meanwhile, step-wise loss and metric trajectories across all training and validation epochs were recorded as TensorBoard logs and archived as run artifacts. Furthermore, best-performing model checkpoints and final evaluation test summaries were automatically archived to allow for exact auditability and comparison across different models.

To guarantee a strictly fair comparison environment, all benchmarking experiments were executed on identical hardware infrastructure using NVIDIA A100-PCIe (40 GB) GPUs running CUDA version 12.9. By integrating every model into the unified AiTLAS framework, we ensured that all architectures were subjected to the exact same data loaders, dataset splits, and preprocessing pipelines. 

\begin{figure}[htbp]
    \centering
    \includegraphics[width=1\textwidth]{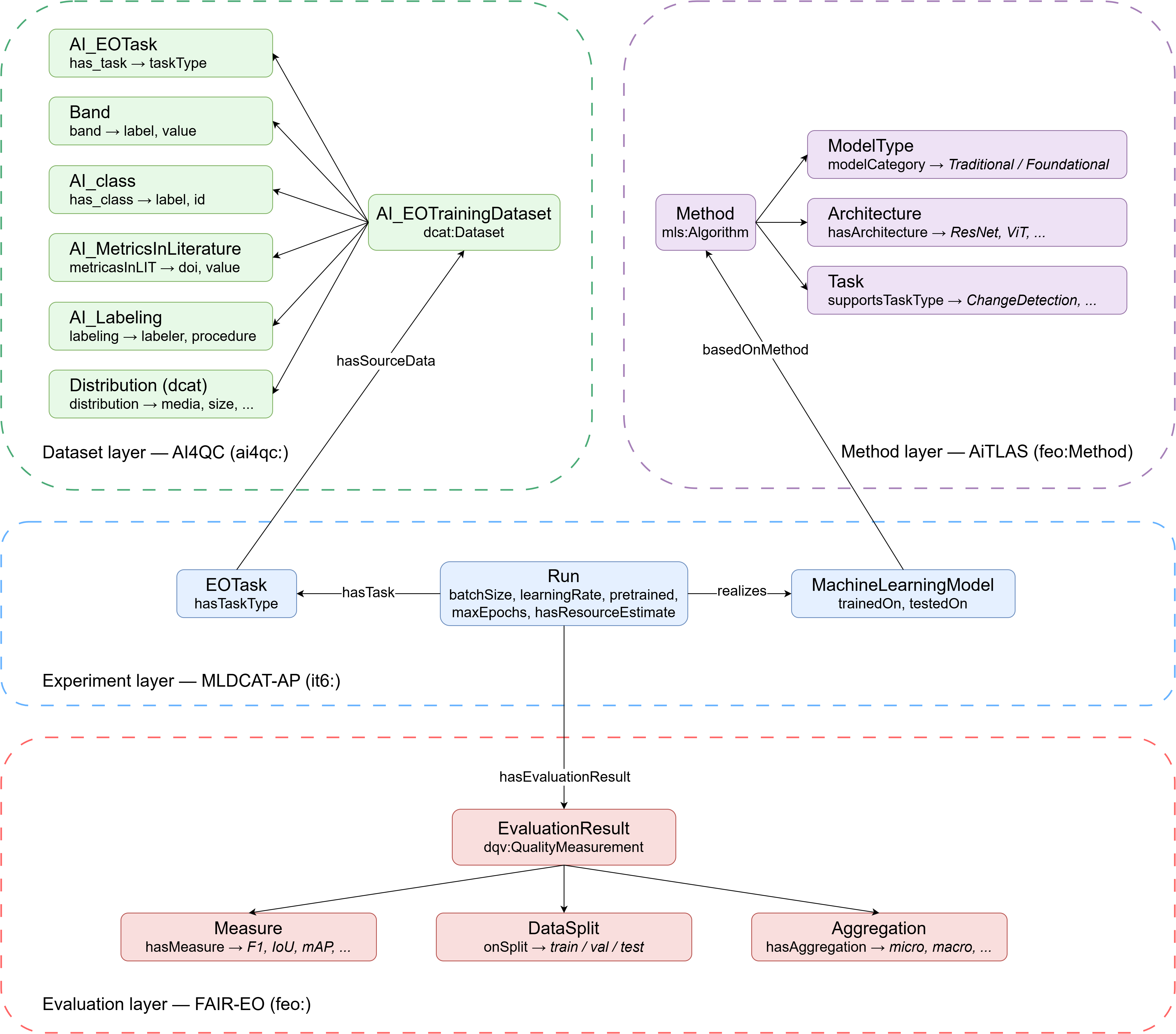}
    \caption{Ontological schema used by the benchmark.}
    \label{fig:schema}
\end{figure}

To maximize the long-term value and auditability of the benchmark, we release its artifacts following the FAIR principles \citep{Wilkinson2016}, applied not only to data but also to the code \citep{Barker2022} and the trained models \citep{Huerta2023}. Concretely, the benchmark comprises four classes of artifacts: datasets and their standardized splits, the benchmarking code, the trained model checkpoints, and the experiment/logs---to each of which the principles apply:

\begin{itemize}
    \item \textbf{Findable:} All artifacts are deposited in a public and Zenodo record with a persistent digital object identifier (DOI) and rich descriptive metadata, and are additionally indexed on GitHub and Hugging Face.
    \item \textbf{Accessible:} Artifacts are openly retrievable on various platforms (GitHub, Hugging Face, Zenodo) under an open license, with metadata remaining available independently of the data payload.
    \item \textbf{Interoperable:} Every model is integrated into the open-source AiTLAS toolbox \citep{Dimitrovski2023, AiTLAS2026} with shared data loaders and standardized split definitions. Beyond code-level interoperability, every dataset, method, run and evaluation result is described in RDF using the ontological schema of Figure~\ref{fig:schema}, which aligns the benchmark with DCAT \citep{dcat} and MLDCAT-AP \citep{mldcat} and makes its results queryable and machine-actionable rather than merely human-readable.
    \item \textbf{Reusable:} Each artifact is accompanied by explicit licensing, provenance recorded through the MLflow experiment logs \citep{Zaharia2018}, and documentation in the form of dataset datasheets \citep{Gebru2021} and model cards \citep{Mitchell2019}, so that individual models, splits, or results can be reused and extended without re-running the full benchmark.
\end{itemize}

To make the benchmark machine-actionable and not merely human-readable, we further describe its datasets, methods, and experiments using an ontological schema that provides a shared, semantically explicit vocabulary for benchmark entities and their relations. This schema underpins an emerging web-based semantic catalogue of EO datasets, methods, and experiments. This change detection benchmark is a first step towards a broad semantic catalogue, and a full description of the catalogue and its ontology is the subject of ongoing work. 

The schema is organized into four layers---Dataset, Method, Experiment and Evaluation---each anchored in a primary vocabulary: the AI4QC ontology (\texttt{ai4qc:}), the FAIR-EO vocabulary (\texttt{feo:}), MLDCAT-AP and the FAIR-EO evaluation terms, respectively. These are cross-referenced with established vocabularies for the concepts they already cover---DCAT (\texttt{dcat:}) for dataset and distribution metadata, ML Schema (\texttt{mls:}) for algorithms, and the W3C Data Quality Vocabulary (\texttt{dqv:}) for quality measurements---yielding a single structure that describes benchmarking experiments end-to-end, from the datasets and methods used through to how each run was executed and evaluated.  The schema, described hereafter, can be seen in Figure ~\ref{fig:schema}. 

Firstly, the Dataset layer describes the training datasets themselves. The latter were semantically annotated using the Artificial Intelligence for Quality Control (AI4QC) ontology \citep{ai4qc_ontology}. Each dataset is linked to a \textit{dcat:Dataset}, connecting this layer to the standard Data Catalog Vocabulary (DCAT) \citep{dcat}. Metadata fields include the EO task the datasets support, the spectral bands involved, class definitions, distribution format, any metrics reported for the dataset in the literature, etc. 

Secondly, the Methods layer describes the models and algorithms being benchmarked, using \textit{feo:Method} as the anchor concept. This includes the model's architecture, model type, and the task it supports. Furthermore, the Experiment layer sits at the center of the schema and ties everything together. A \textit{Task} connects to a \textit{Run}, which records the parameters and resource usage of a specific execution. Each run realizes a \textit{MachineLearningModel}. Two relations carry the cross-layer links: \textit{hasSourceData} connects the \textit{EOTask} to the \textit{AI\_EOTrainingDataset} it was executed on, and \textit{basedOnMethod} connects the resulting \textit{MachineLearningModel} back to the \textit{Method} it instantiates, so that the Experiment layer resolves into the Dataset and Method layers rather than duplicating their
content. This layer is populated using the MLDCAT-AP catalog \citep{mldcat}. 

Finally, the Evaluation layer captures the results of each run as an \textit{EvaluationResult}. Each result specifies the measure used (e.g. IoU), the data split it was computed on (train/val/test), the aggregation method (e.g. micro, macro), and the resulting numeric value. 

Together, these layers make it possible to trace a single evaluation result all the way back to the exact dataset, model configuration, and experimental run that produced it, supporting full reproducibility and structured comparison across the benchmark. All experimental artifacts are available as turtle annotated files on the GitHub repository of the benchmark.

\section{Results \& discussion}
\label{sec:results_discussion}

\subsection{Predictive performance}
\label{sub:performance}

Table~\ref{tab:miou} presents the quantitative evaluation of the benchmarked architectures using mIoU across all ten change detection datasets, comparing models trained from scratch against those initialized with backbones pre-trained on ImageNet-1K. When trained from scratch, SiamCRNN achieves the highest overall performance with an average mIoU of $0.7392$, followed closely by ChangeViT ($0.7381$) and U-Net SiamConc ($0.7224$). When leveraging pre-trained backbones, U-Net SiamConc leads the benchmark with an average mIoU of $0.7931$, followed by SiamCRNN ($0.7797$) and ChangeViT ($0.7709$). Note that weights pre-trained on ImageNet-1K were not available for ChangeFormerV6 and CSSM. Overall, transfer learning yields substantial performance gains across most architectures, raising the benchmark-wide average mIoU from $0.6980$ to $0.7283$.

A key insight from the benchmark is the remarkable resilience and consistent performance of classical convolutional and hybrid architectures. SiamCRNN benefits significantly from combining Siamese convolutional encoders with recurrent layers, securing the highest average mIoU when trained from scratch ($0.7392$) and reaching second-best result ($0.7797$) under pre-training regime while being the best-performing model on MSBC ($0.8950$) and EGY-BCD ($0.8445$) datasets. Meanwhile, the fully convolutional U-Net SiamConc baseline demonstrates exceptional stability. When trained from scratch, it leads CLCD ($0.7048$) and LEVIR-CD+ ($0.7201$). When initialized with pre-trained weights, it emerges as the top-performing model overall ($0.7931$ average mIoU), securing the highest rank across seven datasets including BANDON ($0.6114$), CLCD ($0.7521$), DSIFN ($0.8569$), LEVIR-CD+ ($0.7478$), MSOSCD ($0.8169$), OMBRIA ($0.7114$), and Season-varying CDD ($0.9407$). These results underscore that well-designed convolutional skip-connection mechanisms remain highly competitive against more complex modern architectures.

\begin{table}[htbp]
    \centering
    \caption{Mean Intersection over Union (mIoU) benchmark results across ten change detection datasets, comparing models trained from scratch versus pre-trained backbones. The best performance per dataset within each training setup is indicated in \textbf{bold}, and the second-best is \underline{underlined}.}
    \vspace{5pt}
    \resizebox{\textwidth}{!}{ 
    \begin{tabular}{|l|cccccccccc|c|}
    \hline
    \multicolumn{12}{|c|}{\textbf{(a) Trained from scratch}} \\
    \hline
    \textbf{Dataset} & \rotatebox{90}{\textbf{BIT}} & \rotatebox{90}{\textbf{CGNet}} & \rotatebox{90}{\textbf{ChangeFormerV6}} & \rotatebox{90}{\textbf{ChangeViT}} & \rotatebox{90}{\textbf{CSSM}} & \rotatebox{90}{\textbf{HRNet SiamConc}} & \rotatebox{90}{\textbf{SiamCRNN}} & \rotatebox{90}{\textbf{STANet}} & \rotatebox{90}{\textbf{TinyCD}} & \rotatebox{90}{\textbf{U-Net SiamConc}} & \rotatebox{90}{\textbf{Average}} \\
    \hline
    BANDON & 0.5015 & 0.4785 & \textbf{0.5661} & \underline{0.5545} & 0.4787 & 0.5231 & 0.5511 & 0.4921 & 0.5306 & 0.4840 & 0.5160 \\
    CLCD & 0.6920 & 0.6593 & 0.6569 & 0.6637 & 0.4596 & 0.6375 & \underline{0.6950} & 0.6120 & 0.6612 & \textbf{0.7048} & 0.6442 \\
    DSIFN & 0.7711 & \textbf{0.8109} & 0.7453 & \underline{0.7919} & 0.6829 & 0.6409 & 0.7863 & 0.7224 & 0.7133 & 0.7593 & 0.7424 \\
    EGY-BCD & 0.7783 & \textbf{0.8214} & 0.7594 & 0.7770 & 0.7005 & 0.7619 & \underline{0.7929} & 0.7598 & 0.7619 & 0.7762 & 0.7689 \\
    LEVIR-CD+ & 0.6378 & 0.6412 & 0.6561 & 0.7181 & 0.5785 & 0.6766 & 0.6907 & 0.5449 & \underline{0.7201} & \textbf{0.7201} & 0.6584 \\
    MSBC & 0.8613 & \textbf{0.8846} & \underline{0.8730} & 0.8644 & 0.6554 & 0.8172 & 0.8729 & 0.7875 & 0.7809 & 0.8699 & 0.8267 \\
    MSOSCD & 0.6603 & 0.6748 & 0.6612 & \underline{0.6826} & 0.5330 & 0.6140 & \textbf{0.7248} & 0.6094 & 0.6003 & 0.6817 & 0.6442 \\
    OMBRIA & 0.6438 & 0.6343 & \textbf{0.6934} & \underline{0.6632} & 0.6396 & 0.6123 & 0.6302 & 0.5798 & 0.6138 & 0.6302 & 0.6341 \\
    Season-varying CDD & 0.8896 & 0.6578 & 0.8110 & \textbf{0.9141} & 0.7507 & 0.8584 & \underline{0.9133} & 0.8296 & 0.8556 & 0.8887 & 0.8369 \\
    SYSU-CD & 0.7066 & \textbf{0.7646} & 0.7212 & \underline{0.7512} & 0.6909 & 0.6654 & 0.7350 & 0.6709 & 0.6650 & 0.7090 & 0.7080 \\
    \hline
    \textbf{Average} & 0.7142 & 0.7027 & 0.7144 & \underline{0.7381} & 0.6170 & 0.6807 & \textbf{0.7392} & 0.6608 & 0.6903 & 0.7224 & \\
    \hline
    \hline
    \multicolumn{12}{|c|}{\textbf{(b) Pre-trained}} \\
    \hline
    \textbf{Dataset} & \rotatebox{90}{\textbf{BIT}} & \rotatebox{90}{\textbf{CGNet}} & \rotatebox{90}{\textbf{ChangeFormerV6}} & \rotatebox{90}{\textbf{ChangeViT}} & \rotatebox{90}{\textbf{CSSM}} & \rotatebox{90}{\textbf{HRNet SiamConc}} & \rotatebox{90}{\textbf{SiamCRNN}} & \rotatebox{90}{\textbf{STANet}} & \rotatebox{90}{\textbf{TinyCD}} & \rotatebox{90}{\textbf{U-Net SiamConc}} & \rotatebox{90}{\textbf{Average}} \\
    \hline
    BANDON & 0.4780 & 0.4623 & / & 0.5844 & / & 0.5672 & \underline{0.5957} & 0.4979 & 0.5568 & \textbf{0.6114} & 0.5442 \\
    CLCD & 0.7104 & 0.5061 & / & 0.7101 & / & \underline{0.7182} & 0.7088 & 0.5198 & 0.6821 & \textbf{0.7521} & 0.6634 \\
    DSIFN & 0.7551 & 0.8396 & / & 0.8282 & / & 0.8044 & \underline{0.8459} & 0.8075 & 0.7819 & \textbf{0.8569} & 0.8149 \\
    EGY-BCD & 0.8065 & 0.7409 & / & 0.8066 & / & \underline{0.8433} & \textbf{0.8445} & 0.8074 & 0.7619 & 0.8338 & 0.8056 \\
    LEVIR-CD+ & 0.6690 & 0.5592 & / & \underline{0.7317} & / & 0.7249 & 0.7250 & 0.7031 & 0.7254 & \textbf{0.7478} & 0.6983 \\
    MSBC & 0.8849 & 0.7223 & / & 0.8738 & / & 0.8892 & \textbf{0.8950} & 0.8592 & 0.7762 & \underline{0.8943} & 0.8494 \\
    MSOSCD & 0.7924 & 0.5820 & / & 0.7735 & / & 0.7750 & \underline{0.8076} & 0.7280 & 0.5749 & \textbf{0.8169} & 0.7313 \\
    OMBRIA & 0.6684 & \underline{0.7051} & / & 0.6992 & / & 0.6952 & 0.6903 & 0.6282 & 0.6593 & \textbf{0.7114} & 0.6821 \\
    Season-varying CDD & 0.8946 & 0.7924 & / & 0.9243 & / & 0.9224 & \underline{0.9334} & 0.8964 & 0.8501 & \textbf{0.9407} & 0.8943 \\
    SYSU-CD & 0.7227 & 0.7279 & / & \textbf{0.7774} & / & 0.7166 & 0.7508 & 0.6642 & 0.7298 & \underline{0.7656} & 0.7319 \\
    \hline
    \textbf{Average} & 0.7382 & 0.6638 & / & 0.7709 & / & 0.7657 & \underline{0.7797} & 0.7112 & 0.7098 & \textbf{0.7931} & \\
    \hline
    \end{tabular}
    }
    \label{tab:miou}
\end{table}

Vision transformers also demonstrate strong global feature extraction capabilities across complex datasets. When trained from scratch, ChangeViT achieves the second-highest overall average ($0.7381$), winning Season-varying CDD ($0.9141$) and placing second-best across five datasets, including DSIFN ($0.7919$), OMBRIA ($0.6632$), and SYSU-CD ($0.7512$). With pre-training, ChangeViT reaches an average mIoU of $0.7709$ and takes top rank on SYSU-CD ($0.7774$). Similarly, ChangeFormerV6 demonstrates strong performance, winning BANDON ($0.5661$) and OMBRIA ($0.6934$) when trained from scratch.

The benchmark also highlights notable performance traits among prior-guided, lightweight, and state-space architectures. CGNet demonstrates strong task-specific accuracy when trained from scratch, winning four datasets: DSIFN ($0.8109$), EGY-BCD ($0.8214$), MSBC ($0.8846$), and SYSU-CD ($0.7646$). However, lower scores on complex or off-nadir datasets like BANDON ($0.4785$) reduce its average to $0.7027$. Meanwhile, the ultra-lightweight TinyCD maintains competitive overall averages ($0.6903$ from scratch, $0.7098$ pre-trained), tying for top rank on LEVIR-CD+ from scratch ($0.7201$) and proving that compact models can achieve high accuracy. Conversely, the selective state-space model CSSM yields the lowest overall performance when trained from scratch ($0.6170$), indicating that linear state-space sequence modeling requires further adaptation to isolate subtle change signals in complex Earth observation imagery.

Furthermore, cross-dataset averages emphasize significant variance in baseline task difficulty. Highly structured datasets, such as Season-varying CDD ($0.8369$ from scratch, $0.8943$ pretrained) and MSBC ($0.8267$ from scratch, $0.8494$ pretrained), consistently yield high mIoU scores across almost all models. In contrast, off-nadir aerial views or disaster-response datasets like BANDON ($0.5160$ from scratch, $0.5442$ pretrained) and OMBRIA ($0.6341$ from scratch, $0.6821$ pretrained) present substantially lower performance across the board. This gap highlights remaining open challenges in domain generalization and spatial alignment for automated Earth observation change detection.

\begin{figure}[H]
    \centering
    \includegraphics[width=1\textwidth]{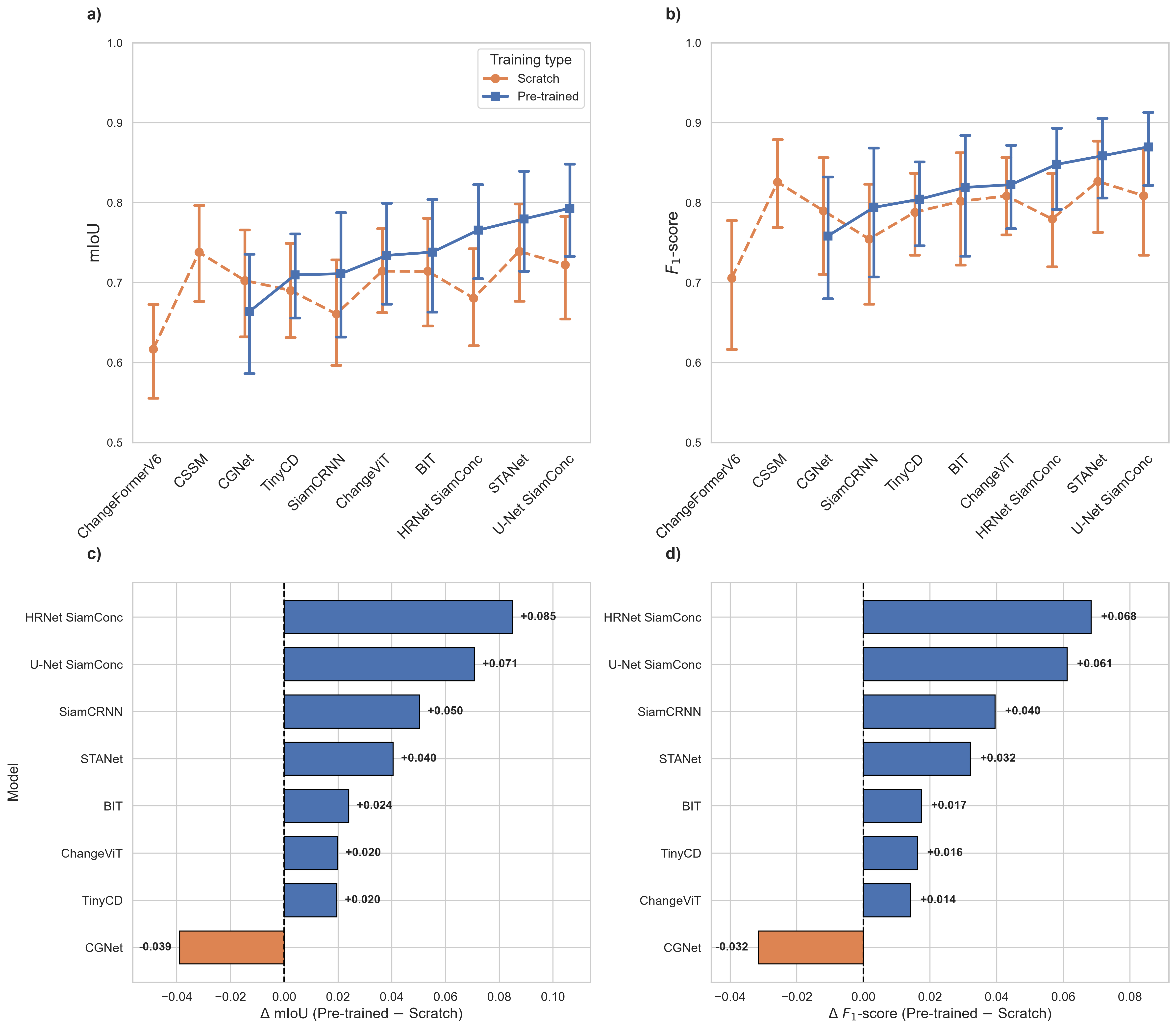}
    \caption{Comparison of a) mIoU and b) macro $F_1$-score for the benchmarked change detection models. Points represent the metric values averaged across the ten benchmark datasets for each architecture. Error bars denote the 95\% confidence interval, indicating the performance variance across the datasets. The models are evaluated under two different training regimes: trained from scratch (orange circles, dashed line) and initialized with pre-trained weights (blue squares, solid line). Figures c) and d) present diverging bar charts illustrating the absolute performance difference ($\Delta$) between the pre-trained models and models trained from scratch. Blue bars extending to the right indicate a performance gain from pre-training, while orange bars to the left indicate a degradation. Models lacking pre-trained weights (ChangeFormerV6 and CSSM) are excluded from the delta comparisons.}
    \label{fig:miou_f1}
\end{figure}

Finally, Fig.~\ref{fig:miou_f1}(a,b) visually shows the trends discussed earlier. In addition to mIoU, it displays the macro $F_1$-score. Notably, there is a strong positive correlation between mIoU and macro $F_1$-score across all evaluated models. Additionally, the wide error bars on both metrics reflect high variance across the datasets. This confirms that while top-tier architectures maintain higher overall baselines, every model remains highly sensitive to the specific complexities and challenges of the target dataset. Note that the data points are intentionally offset along the x-axis to prevent markers and error bars from overlapping.

Importantly, the clear gap between the solid and dashed lines emphasizes the consistent performance boost gained from using pre-trained weights. This is clearly presented in Fig.~\ref{fig:miou_f1}(c,d), which shows the absolute difference between mIoU and $F_1$-score for pre-trained models and the ones trained from scratch. 

We can see that all model architectures gain from using weights pre-trained on ImageNet-1K ($0.02$ to $0.085$ for average mIoU and $0.014$ to $0.068$ for average $F_1$-score), except for CGNet, whose performance degrades due to instability amplified by pre-trained weights. Notably, models such as HRNet SiamConc and U-Net SiamConc exhibit the largest performance improvements when initialized with pre-trained weights. This behavior can be directly attributed to their architectural design, because the backbone represents the vast majority of the model's total parameter count. Consequently, applying ImageNet-1K weights to these models yields a much larger performance boost compared to architectures where the backbone represents a smaller fraction of the network.

\subsection{Computational efficiency}
\label{sub:efficiency}

As detailed in Table~\ref{tab:models}, the benchmarked architectures span a wide computational spectrum in both memory footprint and floating-point operations. The lightweight models such as TinyCD ($0.29~\text{M}$ parameters, $1.46~\text{GFLOPs}$) and CSSM ($2.59~\text{M}$ parameters, $2.78~\text{GFLOPs}$) operate with minimal resources and are suitable for edge deployment. In contrast, heavy transformer- and attention-based networks such as ChangeFormerV6 ($41.03~\text{M}$ parameters, $138.77~\text{GFLOPs}$) and CGNet ($33.68~\text{M}$ parameters, $87.55~\text{GFLOPs}$) require substantially higher computational resources. Occupying a middle ground, models like ChangeViT, HRNet SiamConc and STANet balance moderate parameter capacities with manageable compute costs. 

\begin{table}[!htbp]
  \centering
  \caption{Resource profiling and computational efficiency of the benchmarked models. Peak activation memory is estimated for a training batch of size 16. Latency and throughput are measured on a single GPU with bi-temporal input tensor of size $2 \times 3 \times 256 \times 256$ (batch size 1 for inference). Training times represent the average wall-clock duration to reach convergence across all benchmark datasets.}
  \vspace{5pt}
  \small
  \resizebox{\textwidth}{!}{%
  \begin{tabular}{|l|c|c|c|c|c|}
    \hline
    \textbf{Model} & \textbf{Activation} & \textbf{Latency} & \textbf{Throughput} & \multicolumn{2}{c|}{\textbf{Average train time (h)}} \\
    \cline{5-6}
    & \textbf{memory (GB)} & \textbf{(ms)} & \textbf{(FPS)} & \textbf{Scratch} & \textbf{Pre-trained} \\
    \hline
    BIT & 8.39 & 11.92 & 83.89 & 2.21$\pm$2.57 & 1.72$\pm$2.06 \\
    \hline
    CGNet & 11.51 & 6.02 & 166.11 & 0.93$\pm$1.02 & 0.15$\pm$0.18 \\
    \hline
    ChangeFormerV6 & 14.78 & 22.32 & 44.80 & 1.53$\pm$1.68 & / \\
    \hline
    ChangeViT & 10.40 & 24.47 & 40.87 & 2.12$\pm$2.40 & 1.58$\pm$1.81 \\
    \hline
    CSSM & 1.64 & 60.63 & 16.49 & 1.79$\pm$2.13 & / \\
    \hline
    HRNet SiamConc & 9.91 & 55.13 & 18.14 & 3.25$\pm$4.34 & 2.64$\pm$3.33 \\
    \hline
    SiamCRNN & 6.55 & 9.02 & 110.86 & 3.21$\pm$3.95 & 2.80$\pm$3.95 \\
    \hline
    STANet & 3.01 & 11.02 & 90.74 & 1.62$\pm$1.69 & 1.92$\pm$2.21 \\
    \hline
    TinyCD & 10.32 & 9.30 & 107.52 & 4.13$\pm$4.74 & 5.64$\pm$6.78 \\
    \hline
    U-Net SiamConc & 8.29 & 10.39 & 96.25 & 1.80$\pm$1.96 & 2.33$\pm$2.80 \\
    \hline
  \end{tabular}%
  }
    \label{tab:compute_efficiency}
\end{table}

Moreover, Table~\ref{tab:compute_efficiency} provides the computational resource requirements, runtime speed, and optimization dynamics of the benchmarked models. Peak activation memory varies significantly across models, with the state-space model CSSM requiring a minimal $1.64~\text{GB}$ compared to the high VRAM footprints of heavy transformer architectures like ChangeFormerV6 ($14.78~\text{GB}$) and CGNet ($11.51~\text{GB}$). In terms of inference, CGNet achieves the fastest execution ($6.02~\text{ms}$, $166.11~\text{FPS}$), followed closely by SiamCRNN ($9.02~\text{ms}$) and TinyCD ($9.30~\text{ms}$), whereas CSSM ($60.63~\text{ms}$) and HRNet SiamConc ($55.13~\text{ms}$) exhibit substantially higher latency overheads. 

Importantly, while pre-training speeds up convergence for models with standard backbones like CGNet ($0.15~\text{hours}$) and ChangeViT ($1.58~\text{hours}$), some other models like TinyCD and U-Net SiamConc get little to no benefit from transfer learning.

\begin{figure}[!htbp]
    \centering
    \includegraphics[width=1\textwidth]{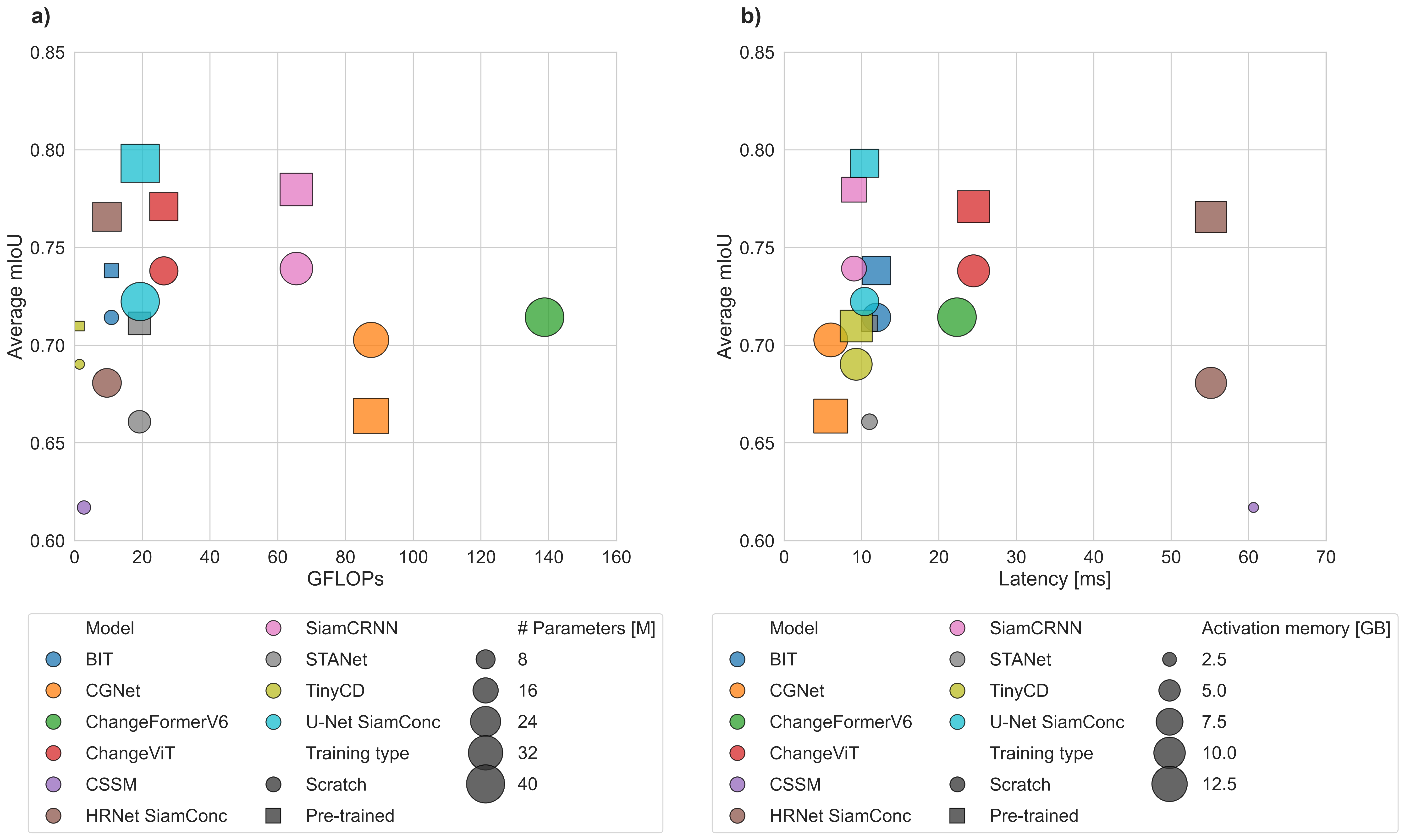}
    \caption{Trade-off between predictive performance and computational efficiency for the benchmarked change detection models. a) Theoretical efficiency displaying average mIoU against GFLOPs, with shape size representing the total number of parameters (in millions). b) Practical hardware efficiency, showcasing average mIoU against inference latency (ms), with bubble size indicating peak activation memory (GB). Marker colors correspond to the specific model architectures, while marker shapes denote the training type: random initialization from scratch (circles) and ImageNet-1K pre-training (squares). Models positioned toward the top-left with smaller shape sizes represent the optimal model architecture, yielding high performance at low resource costs.}
    \label{fig:compute_efficiency}
\end{figure}

Figure~\ref{fig:compute_efficiency} highlights the trade-off between predictive accuracy and computational cost, revealing U-Net SiamConc and SiamCRNN as the most efficient architectures overall. When pre-trained, U-Net SiamConc achieves the highest overall accuracy ($0.7931$ average mIoU) with low latency ($10.39~\text{ms}$) and moderate memory ($8.29~\text{GB}$). SiamCRNN provides a faster ($9.02~\text{ms}$) and more memory-efficient ($6.55~\text{GB}$) alternative while maintaining highly competitive accuracy ($0.7797$ mIoU). These results align with findings by \citet{Corley2024}, confirming that architectures based on U-Net outperform SOTA model architectures from change detection.

At the extremes, CGNet delivers the fastest inference ($6.02~\text{ms}$) but is uniquely hindered by pre-training (performance dropping from $0.7027$ to $0.6638$ mIoU). For strictly memory-constrained deployments, STANet ($3.01~\text{GB}$) provides a highly practical balance, outperforming the lowest-memory but exceptionally slow CSSM ($1.64~\text{GB}$, $60.63~\text{ms}$). Conversely, heavyweight models like ChangeFormerV6, ChangeViT, and HRNet SiamConc incur high computational cost (reaching up to $14.78~\text{GB}$ memory and $55.13~\text{ms}$ latency). While HRNet SiamConc benefits the most from pre-training ($+0.0850$ mIoU), it remains slow for real-time use. Ultimately, pre-training consistently provides an accuracy boost with no additional inference cost. \looseness=-1

\subsection{Qualitative analysis}

\begin{figure}[t]
    \centering
    \includegraphics[width=1\textwidth]{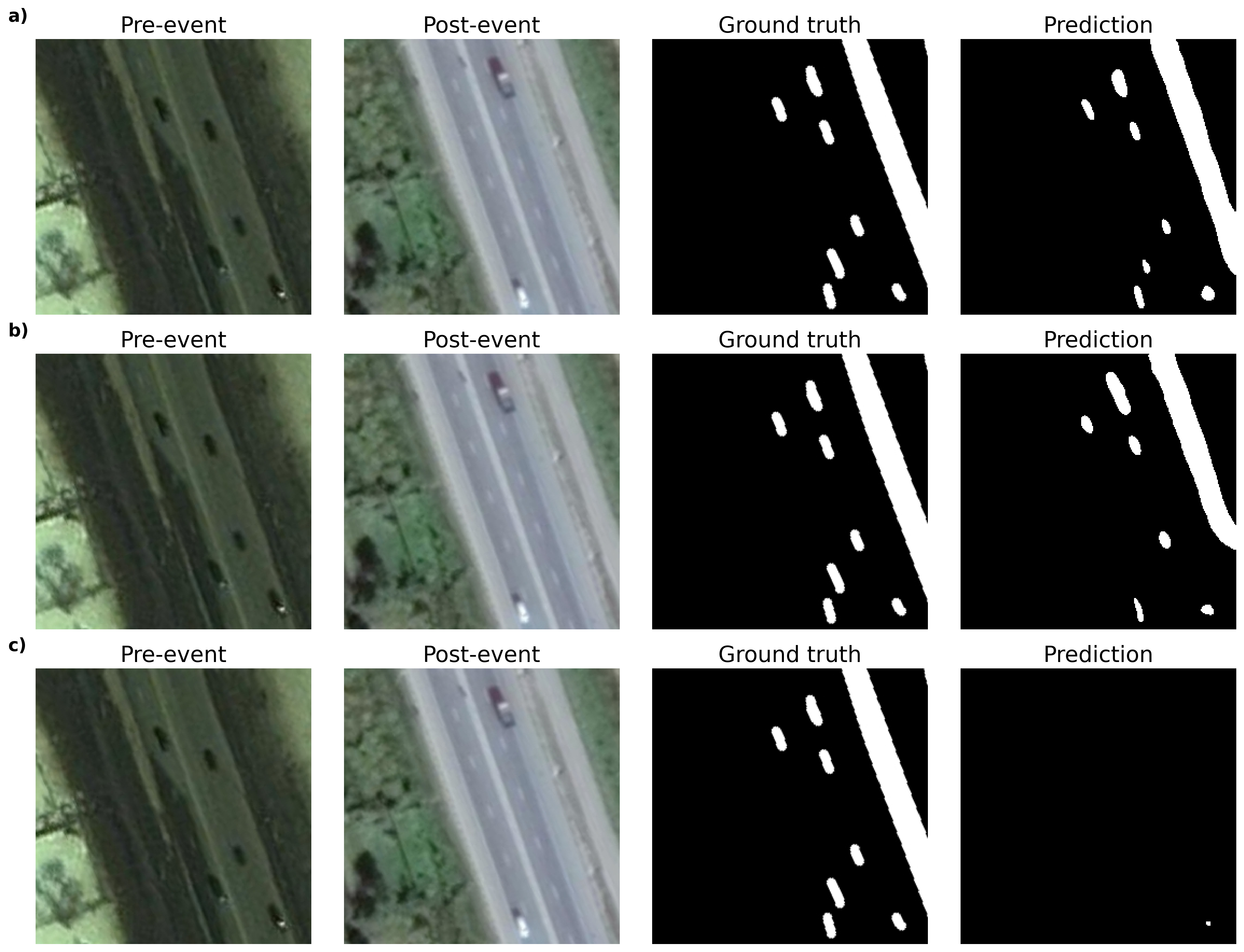}
    \caption{Qualitative comparison of model predictions on the Season-varying CDD dataset. The figure displays the bi-temporal input imagery, ground truth masks, and predicted change maps for three models trained from scratch: a) ChangeViT, b) U-Net SiamConc, and c) CGNet.}
    \label{fig:visual_seasonvaryingcdd}
\end{figure}

\begin{figure}[t]
    \centering
    \includegraphics[width=1\textwidth]{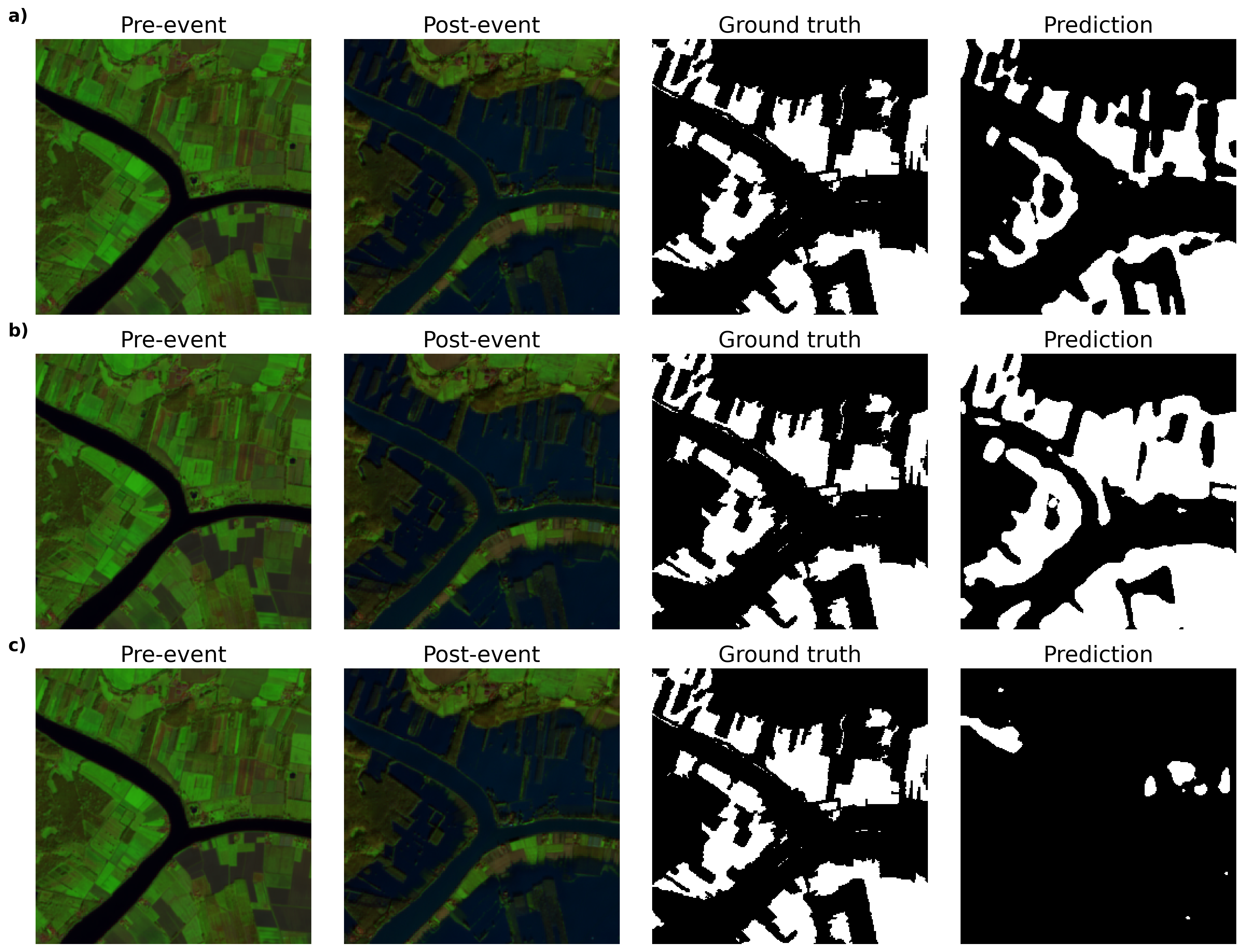}
    \caption{Qualitative comparison of model predictions (flood mapping) on the OMBRIA dataset. The figure displays the bi-temporal input imagery, ground truth masks, and predicted change maps for three models trained from scratch: a) ChangeFormerV6, b) U-Net SiamConc, and c) STANet.}
    \label{fig:visual_ombria}
\end{figure}

To complement the quantitative performance metrics and computational efficiency, we conduct a qualitative visual assessment of the model predictions. Figures \ref{fig:visual_seasonvaryingcdd} and \ref{fig:visual_ombria} present representative examples drawn from the Season-varying CDD and OMBRIA datasets, respectively. Both figures display the bi-temporal input imagery (pre- and post-event), the corresponding ground truth masks, and the predicted change maps generated by selected models. 

This visual comparison allows us to examine the behavioral characteristics of the models that aggregate metrics often obscure, such as their precision in delineating sharp object boundaries, their sensitivity to small-scale structural changes, and their robustness against false positives triggered by seasonal or illumination variations.

Specifically, Fig.~\ref{fig:visual_seasonvaryingcdd} shows the results for a) ChangeViT, b) U-Net SiamConc and c) CGNet trained from scratch on Season-varying CDD dataset. ChangeViT was selected for visualization as the best-performing model ($0.9141$ mIoU) and CGNet as the worst-performing ($0.6578$ mIoU) model on Season-varying CDD dataset, while U-Net SiamConc is the overall winner of the benchmark, achieving mIoU of $0.8887$ on this dataset. While the predicted masks for ChangeViT and U-Net SiamConc exhibit high fidelity and closely align with the ground truth, CGNet completely fails to capture any meaningful spatial patterns.

Similarly, Fig.~\ref{fig:visual_ombria} visualizes the results for the following models trained from scratch on the OMBRIA dataset: a) ChangeFormerV6 (the best-performing model on OMBRIA, $0.6934$ mIoU), U-Net SiamConc (best overall model achieving mIoU of $0.6302$ on OMBRIA), and STANet (the worst model on OMBRIA, $0.5798$ mIoU). Notably, the masks predicted by ChangeFormerV6 closely align with the ground truth, producing spatially smooth boundaries while missing only minor structural details. In contrast, STANet struggles significantly, failing to detect the vast majority of the actual changes.

\subsection{Limitations}

Despite the rigorous standardization of our evaluation protocol, we acknowledge certain experimental limitations. Firstly, to ensure a strictly controlled benchmark, training hyperparameters were kept constant across all architectures rather than being individually optimized for each specific model. While this isolates architectural design as the primary variable, it may disadvantage certain models that are highly sensitive to specific hyperparameter configurations, which could be the case for models like CGNet and STANet. 

Furthermore, due to the substantial computational cost of evaluating ten models across ten distinct datasets, each model-dataset combination was executed as a single training run with two different training regimes. This introduces the inherent risk of stochastic training dynamics, where models might occasionally converge to suboptimal local minima. For instance, we observed that the training process for CGNet can be unstable, occasionally triggering premature early stopping, which likely explains its highly variable performance across different datasets. As a result, the reported metrics include some degree of variance and may not capture the absolute peak potential of every architecture under perfectly optimized, multi-run conditions.

\section{Conclusions}
\label{sec:conclusions}

We present a systematic review and evaluation of modern deep learning architectures applied to Earth observation change detection. Specifically, we introduce a standardized, open-source change detection benchmark and demonstrate its utility through a comprehensive analysis of ten state-of-the-art architectures evaluated across ten heterogeneous EO datasets. By establishing a strictly controlled experimental protocol, we evaluated both models trained from scratch and pre-trained models under identical conditions and hardware constraints. Our evaluation goes beyond standard predictive metrics and includes rigorous profiling of computational efficiency, such as latency, FLOPs, and memory footprints.

This is the most comprehensive and auditable benchamrk of the current state of deep learning approaches for change detection, covering different architectures ranging from classical CNNs to modern ViTs on heterogeneous datasets with diverse scenes. Crucially, our results challenge the prevailing narrative in recent literature and support findings by \citet{Corley2024} that classical, well-optimized convolutional architectures like U-Net SiamConc often provide a superior trade-off between accuracy and computational cost compared to highly complex vision transformers or state-space models.

All important details regarding the study design, metadata, training logs, and best-performing model checkpoints are publicly available. This transparent approach will contribute to more systematic and rigorous experiments in future work and will enable better usability and faster development of genuinely novel approaches. We believe that both this study and the associated repository can serve as a starting point and a guiding design principle for benchmarking machine learning methods for change detection in EO.

More generally, by ensuring these openly available resources adhere strictly to FAIR principles, we aim to bring the AI and EO communities closer together. This fully accessible approach allows researchers to directly reuse our assets, significantly reducing the computational overhead of redundant model training.

The change detection benchmark presented in this paper constitutes the first step towards a broader, comprehensive EO benchmarking suite. Future work will encompass other fundamental EO tasks, including multi-class and multi-label image classification, semantic segmentation, and object detection. By extending our rigorous, FAIR-aligned evaluation protocol, we aim to systematically benchmark a wide array of task-specific model architectures alongside emerging EO foundation models across carefully curated, standardized datasets. Ultimately, this unified benchmarking ecosystem, called \textit{FAIR-EO Hub} (\url{https://eodata.bvlabs.ai/ai4eo}), will serve as a comprehensive, standardized platform for evaluating the next generation of AI models across all fundamental EO applications, with a diverse collection of AI-ready datasets already publicly available.

\section*{CRediT authorship contribution statement}

\textbf{Tadej Tomanič:} Formal analysis, Investigation, Methodology, Software, Validation, Visualization, Writing - original draft, Writing - review \& editing; \textbf{Alice Baudhuin:} Data curation, Software, Visualization, Writing - original draft, Writing - review \& editing. \textbf{Jan Sotošek}: Investigation, Methodology, Resources, Software, Visualization. \textbf{Jure Brence:} Conceptualization, Writing - review \& editing. \textbf{Panče Panov:} Funding acquisition, Project administration, Supervision, Writing - review \& editing. \textbf{Nikola Simidjievski:} Conceptualization, Methodology, Funding acquisition, Supervision, Writing - review \& editing. \textbf{Dragi Kocev:} Conceptualization, Funding acquisition, Supervision, Writing - review \& editing.

\section*{Code and Data Availability}

To ensure complete end-to-end reproducibility, all resources developed for this study have been made publicly accessible. The benchmarking framework is fully open-source and integrated into the AiTLAS toolbox (\url{https://github.com/biasvariancelabs/aitlas}). All standardized data splits, data loaders, training scripts, and annotated turtle files required to reproduce the experiments are available in the dedicated GitHub repository at \url{https://github.com/biasvariancelabs/FAIR-EO-CD-benchmark}, together with corresponding schemas and ontologies. Furthermore, the trained model checkpoints and complete TensorBoard training logs have been deposited on Hugging Face (\url{https://huggingface.co/bvlabs/FAIR-EO-CD-benchmark}) and Zenodo (\url{https://zenodo.org/uploads/21915869}). While we do not host the raw data directly, links to the original datasets are readily available on our GitHub page.

\section*{Declaration of competing interest}

The authors declare that they have no known competing financial interests or personal relationships that could have appeared to influence the work reported in this paper.

\section*{Acknowledgements}

This study was conducted as part of \textit{FAIR-EO - FAIR, Open and AI-Ready Earth Observation Resources} project supported by the Open Science Clusters’ Action for Research and Society (OSCARS) project. The OSCARS project has received funding from the European Commission’s Horizon Europe Research and Innovation programme under grant agreement No.101129751. The FAIR-EO project has been funded through the first Cascading Grants call of the OSCARS project. Furthermore, the authors gratefully acknowledge the computing resources provided by the HPC Vega system at the Institute of Information Science (IZUM), Maribor, Slovenia.

\bibliographystyle{elsarticle-harv} 
\bibliography{references} 

\end{document}